\pdfoutput=1

\documentclass[11pt]{article}

\usepackage[preprint]{acl}

\usepackage{times}
\usepackage{latexsym}
\usepackage{booktabs}

\usepackage[T1]{fontenc}

\usepackage[utf8]{inputenc}

\usepackage{microtype}

\usepackage{inconsolata}
\usepackage{graphicx}
\usepackage{svg}

\usepackage[suppress]{color-edits}
\addauthor{gn}{magenta}
\addauthor{an}{orange}
\addauthor{lt}{blue}

\title{CMULAB: An Open-Source Framework for Training and Deployment of Natural Language Processing Models}

\author{Zaid Sheikh$^1$, Antonios Anastasopoulos$^2$, Shruti Rijhwani$^1$,\\ \textbf{Lindia Tjuatja$^1$, Robbie Jimerson$^3$, Graham Neubig$^1$} \\
  $^1$Carnegie Mellon University, $^2$George Mason University, $^3$Rochester Institute of Technology \\
  \texttt{\{zsheikh,gneubig\}@cs.cmu.edu}
}

\begin{document}
\maketitle
\begin{abstract}
Effectively using Natural Language Processing (NLP) tools in under-resourced languages requires a thorough understanding of the language itself, familiarity with the latest models and training methodologies, and technical expertise to deploy these models. This could present a significant obstacle for language community members and linguists to use NLP tools. This paper introduces the CMU Linguistic Annotation Backend (\href{https://cmulab.dev}{CMULAB}), an open-source framework that simplifies model deployment and continuous human-in-the-loop fine-tuning of NLP models. CMULAB enables users to leverage the power of multilingual models to quickly adapt and extend existing tools for speech recognition, OCR, translation, and syntactic analysis to new languages, even with limited training data. We describe various tools and APIs that are currently available and how developers can easily add new models/functionality to the framework. 
Code is available at \href{https://github.com/neulab/cmulab}{github.com/neulab/cmulab} along with a live demo at \href{https://cmulab.dev}{https://cmulab.dev}\footnote{A short demo video is available at \href{https://youtu.be/t67hDN_hly4}{youtu.be/t67hDN\_hly4}}.
\end{abstract}

\section{Introduction}

With improved methodology and data availability, it is now possible for a skilled NLP practitioner to build somewhat accurate systems for many languages in the world, even those that have relatively few resources \citep{hedderich-etal-2021-survey,ogueji-etal-2021-small}.
At the same time, there is burgeoning interest from linguists and language community members to apply these state-of-the-art techniques to languages where these technologies are not widely accessible, for applications such as optical character recognition \citep{nguyen2021survey,rijhwani-etal-2020-ocr}, speech recognition \citep{bartelds-etal-2023-making,Hou2021ExploitingAF}, and machine translation \citep{ranathunga2023neural,Haddow2021SurveyOL}.
However, herein lies a skill gap: despite this considerable interest, often the barrier of entry for building high-quality systems is high, because of the technological expertise that is typically necessary. 

To address this issue, we present the CMU Linguistic Annotation Backend (CMULAB), an open-source web-based framework that allows users to quickly adapt/extend existing NLP tools to new languages and domains by leveraging massively multilingual neural network models.
This allows language community members, linguists, and other end-users to use these models as well as improve them by uploading corrected annotations and fine-tuning the models (\autoref{sec:framework}).

Several NLP tasks are included out-of-the-box (\autoref{sec:nlp-tasks}). These include
\begin{itemize}
\item \textbf{Optical character recognition}, using the Google Cloud Vision OCR\footnote{\url{https://cloud.google.com/vision/docs/ocr}} and the post-correction method of \citet{rijhwani-etal-2020-ocr}.
\item \textbf{Speech recognition}, through Allosaurus, a pretrained universal phone recognizer for more than 2000 languages \cite{li2020universal}.
\item \textbf{Speaker diarization}, through a model based on Resemblyzer\footnote{\url{https://github.com/resemble-ai/Resemblyzer}}, an open-source diarization tool based on neural speaker embeddings~\cite{wan2018generalized}.
\item \textbf{Machine translation}, through the NLLB translation model that supports 200 languages \cite{nllb2022}.
\item \textbf{Morphosyntactic analysis}, with GlossLM, a pre-trained model for interlinear glossing in many languages \cite{ginn2024glosslm}.
\end{itemize}

End-users do not need to have any coding or technical expertise to use or fine-tune these models. In addition, the open-source and modular nature of the framework allows developers to easily integrate additional models or functionality. Developers also have direct access to the back-end models via well-defined REST APIs, allowing them to design custom interfaces or add additional features to popular annotation tools such as FLEx\footnote{https://software.sil.org/fieldworks/}, ELAN \citep{wittenburg-etal-2006-elan} or Inception \citep{klie2018inception}.

CMULAB takes a step towards putting advanced language technologies in the hands of under-represented language communities and linguists, and in \autoref{sec:case-study}, we discuss some of our initial case studies where CMULAB is being used to develop NLP models for endangered languages.

\section{Framework Overview}
\label{sec:framework}

When users interact with CMULAB, they will do so through a frontend, which can be either the \url{https://cmulab.dev} web site, a front-end plugin to annotation software such as ELAN, or (for developers) through REST APIs (\autoref{sec:frontend}). 
Requests originating from this frontend are forwarded to a separate, high-powered backend server for processing (\autoref{sec:backend}).

Out-of-the-box, CMULAB includes a selection of pre-trained multilingual base models that support a number of NLP tasks (\autoref{sec:nlp-tasks}) over hundreds of languages.
All of these models support zero-shot application to new languages, so even for languages that are not covered in the training data, users can get first results on a new language within minutes.

Users also have the option to upload training data to fine-tune the models to further improve performance beyond what is available zero-shot.
After models are trained, users can interact with them through the interface. They have options to view the model details, run predictions, or delete the model if it is no longer needed. To foster collaboration, users also have the option of sharing their trained models with the community by marking them as public.

\begin{table*}[ht!]
\centering
\caption{Languages supported}
\label{tab:lang-support}
\begin{tabular}{l|l|c}
\toprule
Task                & Model                               & Supported Languages \\ \midrule
Phoneme recognition & Allosaurus                          & 2000+               \\ 
Speaker Diarization & Resemblyzer                         & Language-independent                  \\ 
Automatic glossing  & lecslab/glosslm                     & 1800                \\ 
OCR post-correction & Google Vision API + post-correction model             & 200+ languages, 32 scripts                  \\ 
Machine Translation & facebook/nllb-200-distilled-600M    & 200                 \\ \bottomrule
\end{tabular}
\end{table*}

\section{Currently Supported NLP tasks}
\label{sec:nlp-tasks}

CMULAB supports a variety of NLP tasks with the goal of making linguistic documentation easier, including speaker segmentation and diarization, phoneme recognition, automatic glossing, machine translation and OCR post-correction. \autoref{tab:lang-support} provides an overview of the models used and the languages supported. These tasks are described in more detail in the following sections.

\subsection{OCR post-correction}
\label{sec:ocr-post-correction}
 Pre-trained OCR software can struggle with low-resource languages, often producing inaccurate results with high error rates. Our OCR-post-correction tool \citep{rijhwani-etal-2020-ocr} addresses this issue by automatically correcting errors, given a small amount of training data. The process begins with the user uploading a set of document images and receiving transcribed output from the Google Vision API. Alternatively, users may opt to utilize any other OCR tool for the initial transcription. In the event of errors in the output, users can manually correct a selection of files, which can then be uploaded to train a new post-correction model in step 2. This updated model can subsequently be utilized to correct additional first-pass OCR output in step 3. Steps 2 and 3 can be repeated multiple times to progressively improve the model's performance.

\subsection{Phoneme recognition}
\label{sec:elan}

CMULAB's phone transcription API is based on the Allosaurus \citep{li2020universal} pre-trained universal phone recognizer which supports more than 2000 languages. The API also allows users to fine-tune the Allosaurus pre-trained models by uploading a small amount of training data. Our ELAN extension (\autoref{sec:frontend}) takes advantage of these APIs to enable users to generate initial phoneme transcriptions, make corrections to a few examples, and upload them to fine-tune the Allosaurus model. This can be repeated multiple times to iteratively improve the models. 

\subsection{Speaker Diarization}
Our speaker diarization tool is based on the Resemblyzer python package. It uses a pre-trained neural speech encoder~\cite{wan2018generalized}. In particular, the user needs to provide a minimal amount of speaker annotation (a few seconds per speaker), which are used to create an average speaker embedding. The rest of the audio is then diarized based on its similarity to these speaker embeddings. This approach strikes a good balance between performance (with competent accuracy fairly close to other state-of-the-art but also more complicated models) with minimal annotation requirements, while also being computationally efficient.

\subsection{Machine translation}
\label{sec:machine-translation}
We also offer a translation service powered by Meta's NLLB model \citep{nllb2022}, supporting over 200 languages. This service allows users to translate text quickly and easily. Users can also upload their own data to fine-tune the model for their specific needs, potentially improving translation accuracy for their domain or language pair.

\subsection{Interlinear glossing}
\label{sec:interlinear-glossing}

Interlinear glossed text (IGT) is a widely-used format for linguistic annotation, and is an information-dense resource often created by linguists in language documentation and fieldwork settings. IGT is a multi-line format that presents (1) a transcription in the source language, (2) morpheme-aligned labels (glosses) indicating each morpheme's meaning and/or grammatical function, and often (3) a translation in a widely spoken language, such as English. Creating IGT corpora is a time-intensive task, requiring complex morphological analysis and extensive knowledge of the language grammar. To assist linguists in this task, we provide easy access to GlossLM~\cite{ginn2024glosslm}, a model that continually pretrained ByT5~\cite{xue-etal-2022-byt5} on the task of generating the gloss labels given a transcription and optional translation across 1.8k languages.

\begin{figure*}[ht!]
    \centering
    \includegraphics[width=0.85\linewidth]{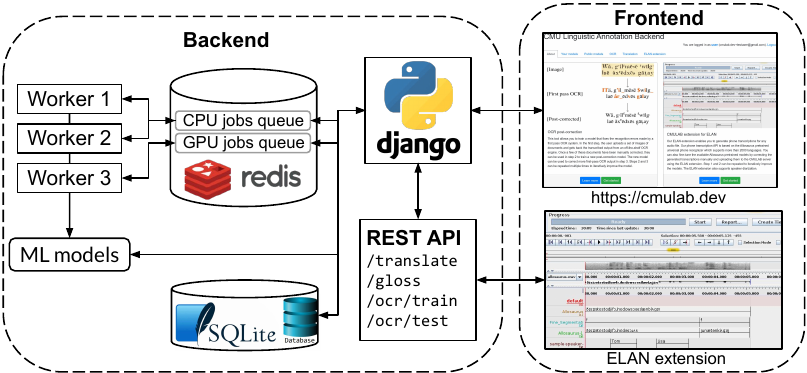}
    \label{fig:backend-architecture}
    \caption{Architecture Diagram}
\end{figure*}

\section{Backend implementation}
\label{sec:backend}

CMULAB's web interface and REST APIs are powered by Django, a popular open-source, Python-based web framework. Backend NLP models and functionality are implemented as python packages that register themselves as CMULAB plugins, enabling automatic discovery. The core CMULAB framework efficiently manages input and output data, invokes registered functions, and routes them to the appropriate task queues. It also supports real-time log file collection and display, task cancellation and restart on user request, email notifications regarding task status and other features to help users monitor and manage their training jobs.

When tasks are submitted, via the web interface or the REST APIs, they are categorized based on their expected resource requirements and placed into corresponding task queues. These task queues are implemented using Redis as the message broker. Multiple background workers constantly fetch and execute tasks from these queues. For horizontal scaling, these workers can be distributed across multiple compute nodes. Pre-built Docker images simplify spinning up new worker nodes on demand. Developers also have the option to deploy their model using a third-party framework and instruct CMULAB to forward requests to the external server.

Authentication is done using Django's built-in authentication system. Third-party OAuth authentication is also supported. API access is enabled using token-based authentication.

\section{Frontend implementation}
\label{sec:frontend}
\begin{figure*}
    \centering
    \includegraphics[width=0.9\linewidth]{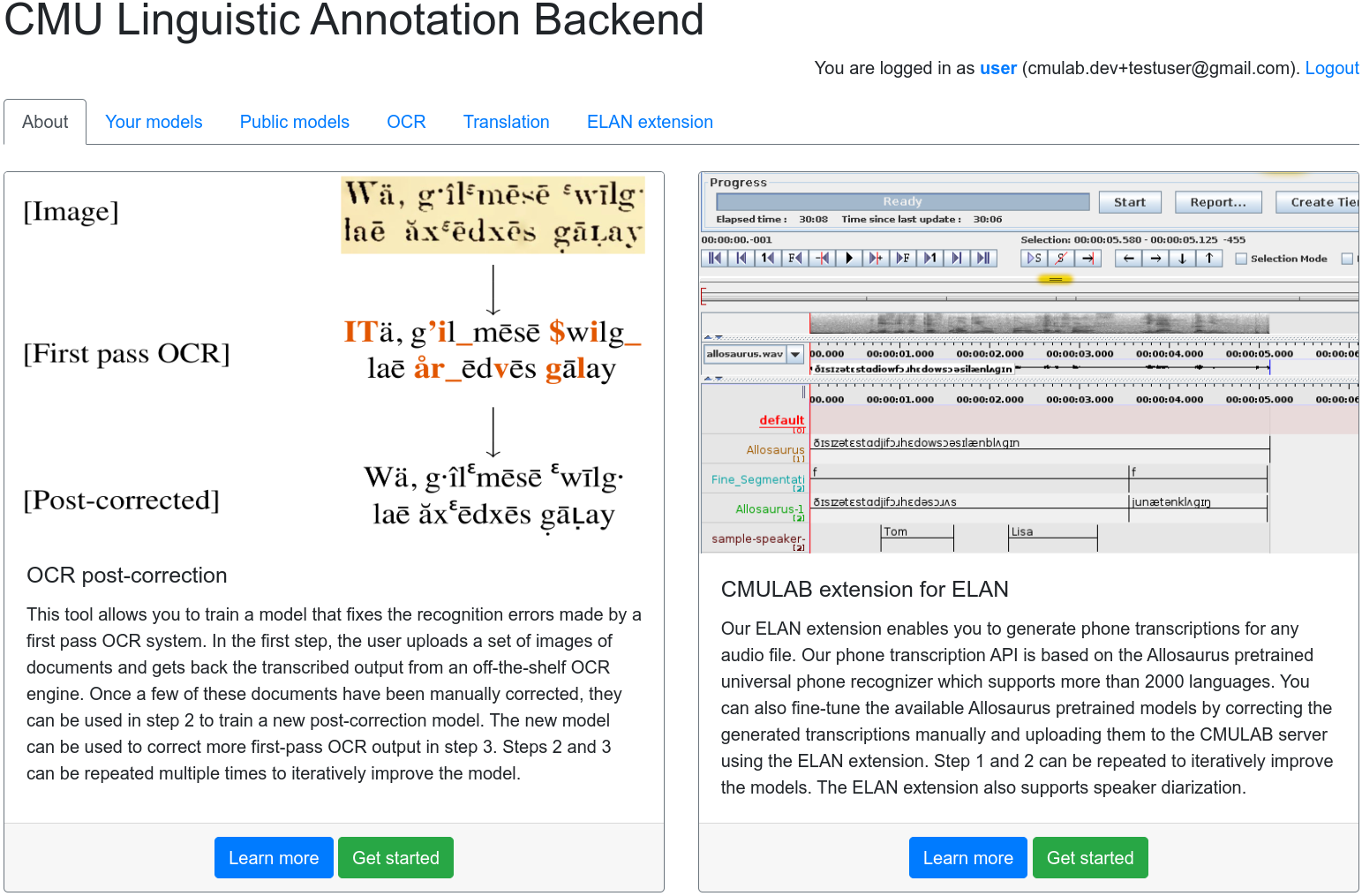}

    \caption{CMULAB Homepage}
\end{figure*}

\begin{figure*}
    \centering
    \includegraphics[width=0.85\linewidth]{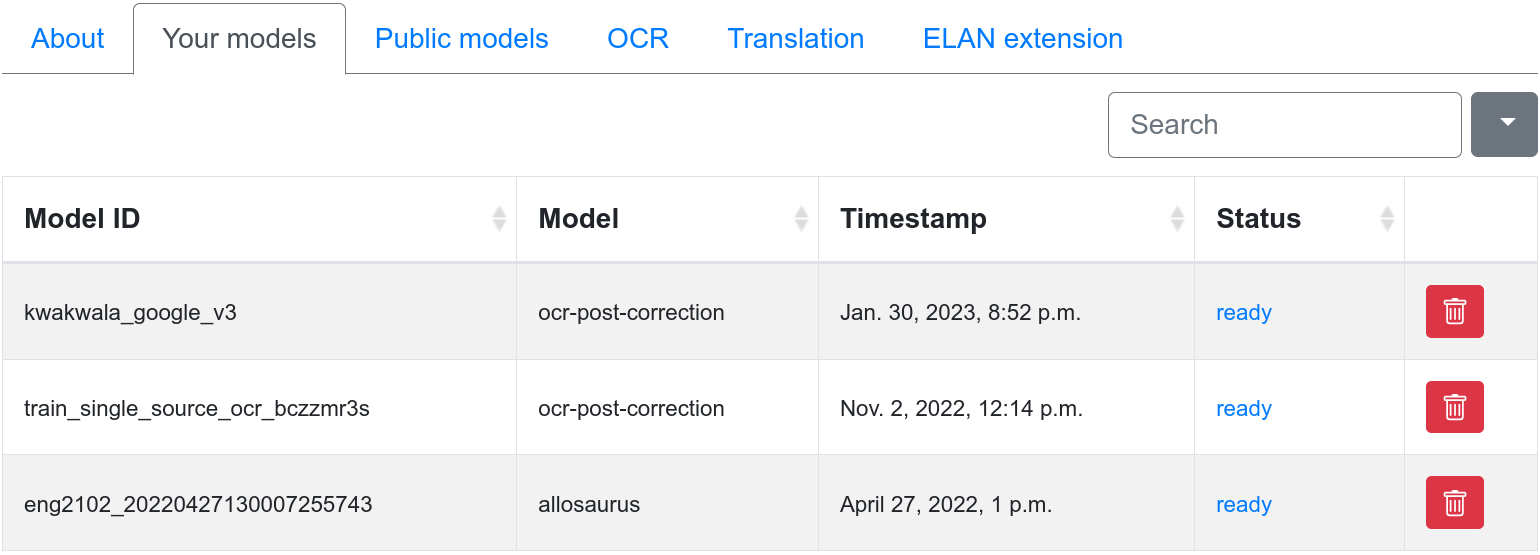}
    \caption{CMULAB Models page}
\end{figure*}

\begin{figure*}
    \centering
    \includegraphics[width=0.85\linewidth]{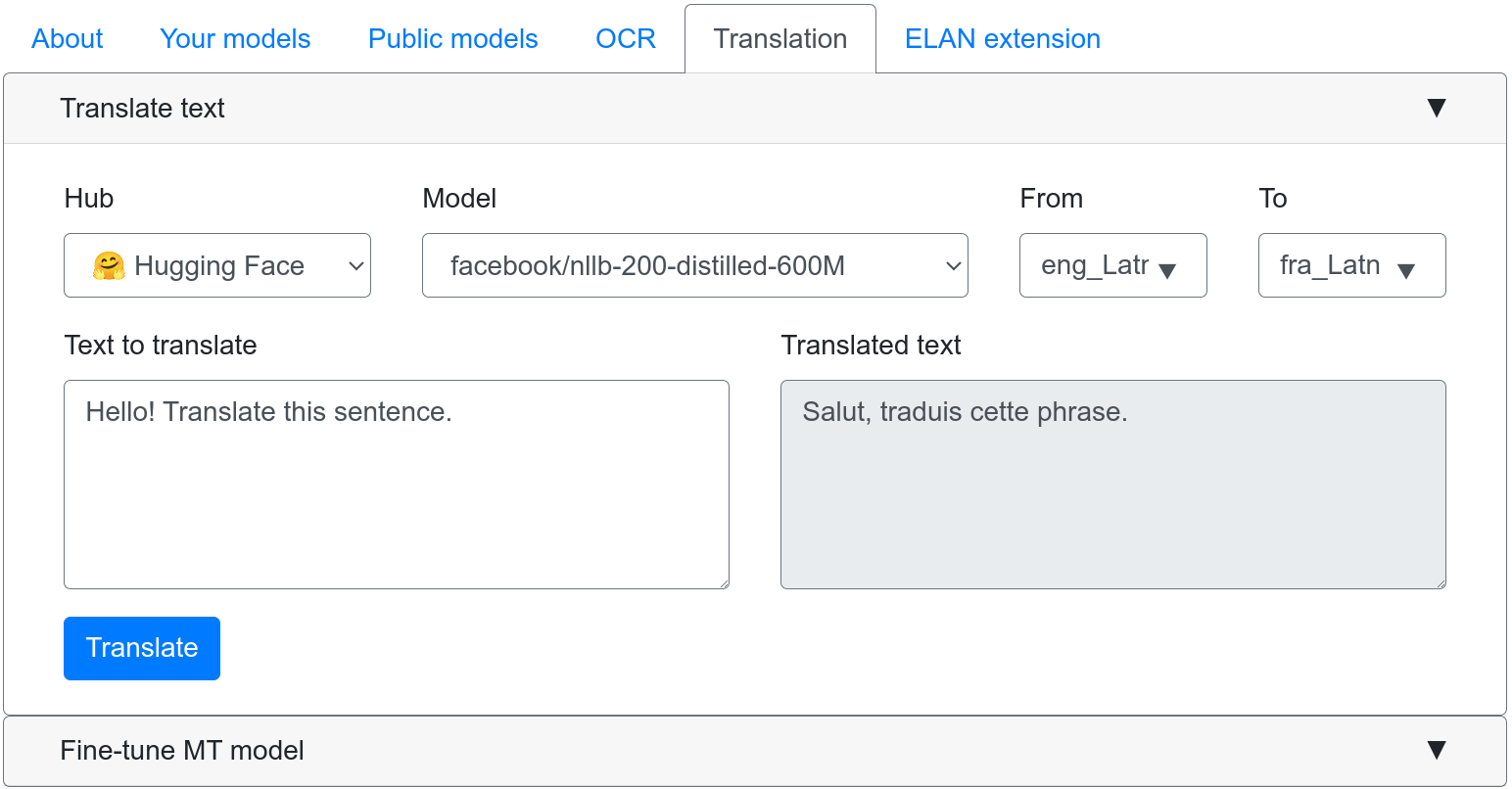}
    \caption{Machine Translation UI}
    \label{fig:mt-ui}
\end{figure*}

\begin{figure*}
    \centering
    \includegraphics[width=0.85\linewidth]{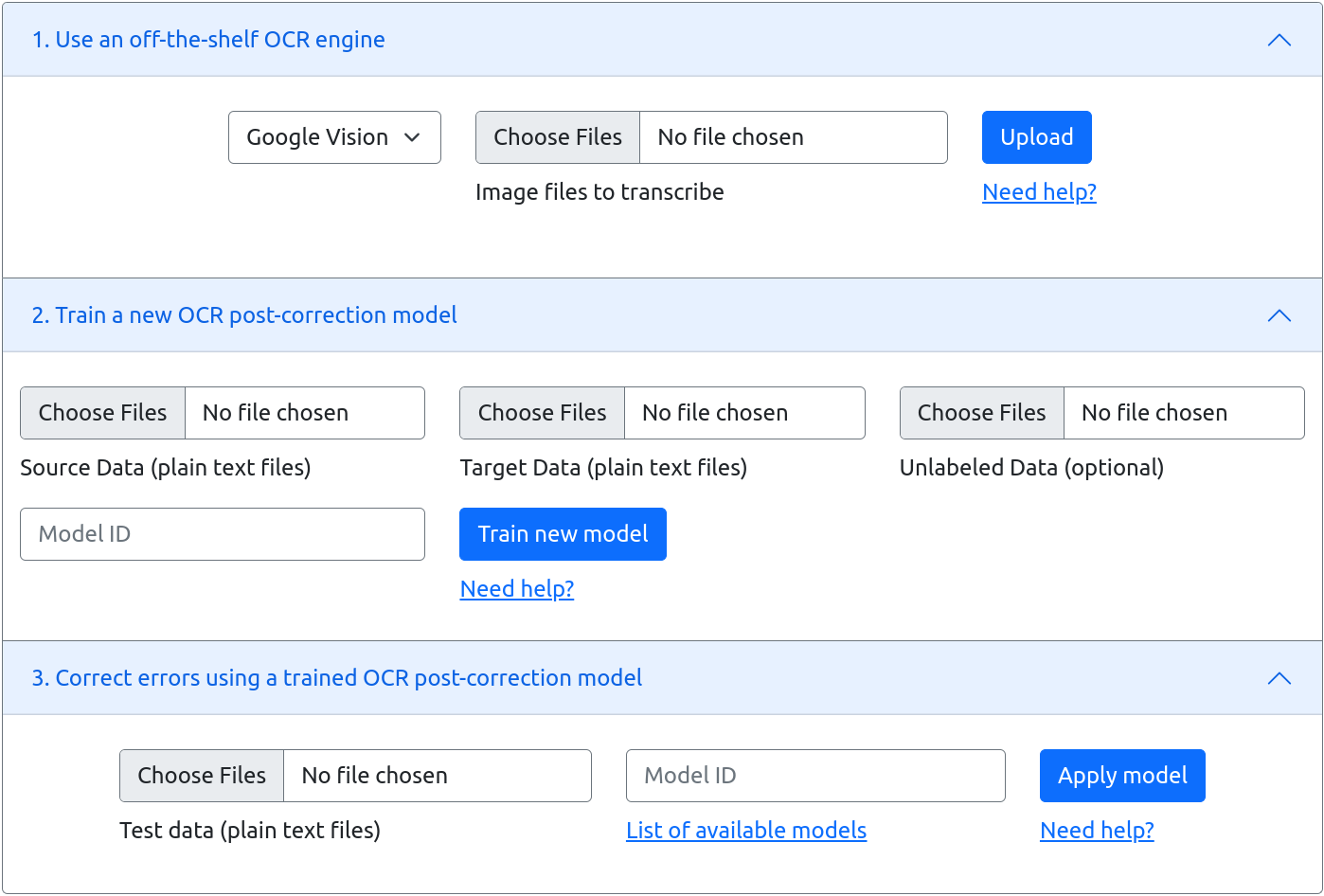}
    \caption{CMULAB OCR post-correction tool}
\end{figure*}

The CMULAB web interface is designed with a focus on maximizing user-friendliness and intuitiveness. Django templates are used server-side for dynamic HTML generation while the client-side functionality is handled by React, a popular JavaScript framework.  Bootstrap CSS is utilized for styling purposes.

Users can also interact with CMULAB through our extension for ELAN,%
\footnote{https://archive.mpi.nl/tla/elan}
a widely-used annotation tool for audio and video recordings. The CMULAB ELAN extension\footnote{https://github.com/neulab/cmulab\_elan\_extension} allows users to run speaker diarization on any audio file, generate phone transcriptions, as well as upload corrected transcriptions to fine-tune the model.

\section{Data Collection and Security}
\label{sec:data-privacy}
All data collected by the main \url{https://cmulab.dev} server is done under an IRB protocol with informed consent. Users can always delete their models and all associated data, including logs. By default, everything remains private. Even if users choose to share a model publicly, only the model itself becomes accessible, not the underlying training data or logs. We also transparently disclose any use of third-party services, such as Google Vision API. In addition, because all of the code is open-source, technically saavy users can set up their own instances on a private server if they wish.

\section{Case Study}
\label{sec:case-study}
As described in \autoref{sec:ocr-post-correction}, CMULAB's OCR post-correction tool allows users to train a model that can correct errors in the output of existing OCR tools. We evaluated its performance and usability using Seneca, a low-resource indigenous language from the Iroquoian family of languages. The data originates from the four books of the gospel translated into Seneca by Presbyterian missionary Asher Wright. Published in 1878 using a printing press at the Seneca Mission on the Buffalo Creek Reservation in Buffalo, New York, the document predates modern computing and was never digitized, remaining solely in PDF format. The initial output from Google Vision API for Seneca data yielded a high CER of 44.11\%. We manually corrected 10 pages and uploaded them to CMULAB. These were used to train a new post-correction model. This achieved a significant reduction in CER, dropping the error rate to 18.53\%.

\section{Comparison to Other Tools}
\label{sec:related-work}
Open-source packages such as Gradio \citep{abid2019gradio} and Streamlit\footnote{https://github.com/streamlit/streamlit} make it easy to quickly build web applications on top of ML models. These frameworks, combined with application packaging and deployment tools such as Replicate's Cog\footnote{https://github.com/replicate/cog} or BentoML\footnote{https://github.com/bentoml/BentoML} allow NLP practitioners to quickly build new NLP tools. However, these frameworks require technical knowledge beyond that available to most linguists or language community members.

On the other end of the spectrum are no-code or low-code open-source GUI frameworks such as H2O LLM Studio \citep{candel-etal-2023-h2o} and LLaMA Board\footnote{https://github.com/hiyouga/LLaMA-Factory} that help users fine-tune LLMs. Deploying them in a scalable manner can still require a lot of technical knowledge and computational resources. The Prompt2Model project \citep{prompt2model} attempts to address this issue by taking a natural language task description and using it to train a special-purpose model that is conducive to deployment. However, these approaches, due to their reliance on LLMs, do not work well on low-resource languages.

CMULAB, on the other hand, focuses on simplifying the use of NLP tools for low-resource languages. To this end, it offers pre-trained multilingual models that support hundreds of languages, a user-friendly interface for non-technical users, and is fully open source with clear contribution paths for researchers and developers to add new features and functionality. 

\section{Conclusion and Future Work}
\label{sec:conclusion}
CMULAB presents a significant step towards democratizing access to NLP tools and models, particularly for under-resourced languages. By offering pre-trained multilingual models, a user-friendly interface, and easy deployment options, CMULAB empowers language communities and linguists to leverage advanced NLP technologies without requiring extensive technical expertise.

While this project has already achieved substantial progress, we envision several avenues for future development. Integrating active learning algorithms could optimize data annotation by prioritizing the most informative data points. Additionally, incorporating model comparison and evaluation tools would allow users to assess and select the best models for their specific needs. We also plan to implement a detailed version history system for uploaded data, enabling users to track changes in model performance over time. Finally, we aim to introduce granular permissions and access control mechanisms to facilitate data sharing, joint data curation, and enhanced collaboration.

\section*{Limitations}
\label{sec:limitations}
While CMULAB offers a powerful, flexible, and accessible platform to democratize access to NLP tools and models, it is important to acknowledge some limitations. CMULAB allows fine-tuning of pre-trained models, but the degree of customization is limited since end-users cannot modify the underlying model architecture or training algorithms. Furthermore, the success of fine-tuning is highly dependent on the quality and quantity of user-provided training data. In addition, CMULAB currently supports a limited set of NLP tasks. While its focus on extensibility allows for easy integration of new functionalities, a certain level of programming skills and knowledge of NLP techniques is still required. This necessitates cooperation between language communities and NLP researchers to build these new tasks.

\section*{Acknowledgements}
We thank numerous collaborators who contributed to usability testing of CMULAB. Daisy Rosenblum and Olivia Chen from the University of British Columbia evaluated our OCR post-correction tool on Kwak'wala data and provided comprehensive feedback. We are also grateful to Jonathan Amith from Gettysburg College, who extensively tested our ELAN plugin using Yoloxochitl Mixtec and Nahuatl recordings, providing valuable suggestions for improvement. 
We also thank Alexis Michaud for his feedback and for suggesting the name CMULAB.
This material is based upon work supported in part by the National Science Foundation under Grants No. 1761548, 2040926, 2211951, and by the National Endowment for the Humanities under Grant \#PR-276810-21.

\section*{Ethics Statement}
The CMULAB framework has the potential to democratize NLP by lowering the entry barrier for researchers and domain specialists who lack extensive technical expertise. This could lead to a wider range of languages and domains being covered by NLP tools, fostering broader understanding and communication across cultures. 

However, with the ease of access and modification of NLP models, several ethical considerations must be taken into account. Multilingual large language models have been shown to facilitate the transmission of biases across different cultures and languages. Bearing this in mind, we intend to introduce measures to detect and mitigate such biases.

In addition, since CMULAB allows users to upload corrections and annotations, privacy and security of the data is of utmost importance, especially when it involves sensitive or personal data. CMULAB prioritizes the protection of user data. All data uploaded to CMULAB remains private by default, and users have full control over its access and deletion. Informed consent is obtained from individuals whose data is used to train or fine-tune NLP models. 

We also acknowledge the importance of respecting intellectual property rights, especially when working with data and models for under-resourced languages and encourage users to be mindful of existing intellectual property rights and to engage in open and collaborative practices that respect the ownership and control of language data.

\nocite{*}

\bibliography{custom}

\begin{thebibliography}{22}
\expandafter\ifx\csname natexlab\endcsname\relax\def\natexlab#1{#1}\fi

\bibitem[{Abid et~al.(2019)Abid, Abdalla, Abid, Khan, Alfozan, and Zou}]{abid2019gradio}
Abubakar Abid, Ali Abdalla, Ali Abid, Dawood Khan, Abdulrahman Alfozan, and James Zou. 2019.
\newblock Gradio: Hassle-free sharing and testing of ml models in the wild.
\newblock \emph{arXiv preprint arXiv:1906.02569}.

\bibitem[{Bartelds et~al.(2023)Bartelds, San, McDonnell, Jurafsky, and Wieling}]{bartelds-etal-2023-making}
Martijn Bartelds, Nay San, Bradley McDonnell, Dan Jurafsky, and Martijn Wieling. 2023.
\newblock \href {https://doi.org/10.18653/v1/2023.acl-long.42} {Making more of little data: Improving low-resource automatic speech recognition using data augmentation}.
\newblock In \emph{Proceedings of the 61st Annual Meeting of the Association for Computational Linguistics (Volume 1: Long Papers)}, pages 715--729, Toronto, Canada. Association for Computational Linguistics.

\bibitem[{Candel et~al.(2023)Candel, McKinney, Singer, Pfeiffer, Jeblick, Lee, and Conde}]{candel-etal-2023-h2o}
Arno Candel, Jon McKinney, Philipp Singer, Pascal Pfeiffer, Maximilian Jeblick, Chun~Ming Lee, and Marcos Conde. 2023.
\newblock \href {https://doi.org/10.18653/v1/2023.emnlp-demo.6} {{H}2{O} open ecosystem for state-of-the-art large language models}.
\newblock In \emph{Proceedings of the 2023 Conference on Empirical Methods in Natural Language Processing: System Demonstrations}, pages 82--89, Singapore. Association for Computational Linguistics.

\bibitem[{Ginn et~al.(2024)Ginn, Tjuatja, He, Rice, Neubig, Palmer, and Levin}]{ginn2024glosslm}
Michael Ginn, Lindia Tjuatja, Taiqi He, Enora Rice, Graham Neubig, Alexis Palmer, and Lori Levin. 2024.
\newblock \href {http://arxiv.org/abs/2403.06399} {Glosslm: Multilingual pretraining for low-resource interlinear glossing}.

\bibitem[{Haddow et~al.(2021)Haddow, Bawden, Barone, Helcl, and Birch}]{Haddow2021SurveyOL}
Barry Haddow, Rachel Bawden, Antonio Valerio~Miceli Barone, Jindvrich Helcl, and Alexandra Birch. 2021.
\newblock \href {https://api.semanticscholar.org/CorpusID:237371890} {Survey of low-resource machine translation}.
\newblock \emph{Computational Linguistics}, 48:673--732.

\bibitem[{Hedderich et~al.(2021)Hedderich, Lange, Adel, Str{\"o}tgen, and Klakow}]{hedderich-etal-2021-survey}
Michael~A. Hedderich, Lukas Lange, Heike Adel, Jannik Str{\"o}tgen, and Dietrich Klakow. 2021.
\newblock \href {https://doi.org/10.18653/v1/2021.naacl-main.201} {A survey on recent approaches for natural language processing in low-resource scenarios}.
\newblock In \emph{Proceedings of the 2021 Conference of the North American Chapter of the Association for Computational Linguistics: Human Language Technologies}, pages 2545--2568, Online. Association for Computational Linguistics.

\bibitem[{Hou et~al.(2021)Hou, Zhu, Wang, Wang, Qin, Xu, and Shinozaki}]{Hou2021ExploitingAF}
Wenxin Hou, Hanlin Zhu, Yidong Wang, Jindong Wang, Tao Qin, Renjun Xu, and Takahiro Shinozaki. 2021.
\newblock \href {https://api.semanticscholar.org/CorpusID:235187234} {Exploiting adapters for cross-lingual low-resource speech recognition}.
\newblock \emph{IEEE/ACM Transactions on Audio, Speech, and Language Processing}, 30:317--329.

\bibitem[{Klie et~al.(2018)Klie, Bugert, Boullosa, de~Castilho, and Gurevych}]{klie2018inception}
Jan-Christoph Klie, Michael Bugert, Beto Boullosa, Richard~Eckart de~Castilho, and Iryna Gurevych. 2018.
\newblock The inception platform: Machine-assisted and knowledge-oriented interactive annotation.
\newblock In \emph{proceedings of the 27th international conference on computational linguistics: system demonstrations}, pages 5--9.

\bibitem[{Li et~al.(2020)Li, Dalmia, Li, Lee, Littell, Yao, Anastasopoulos, Mortensen, Neubig, Black, and Florian}]{li2020universal}
Xinjian Li, Siddharth Dalmia, Juncheng Li, Matthew Lee, Patrick Littell, Jiali Yao, Antonios Anastasopoulos, David~R Mortensen, Graham Neubig, Alan~W Black, and Metze Florian. 2020.
\newblock Universal phone recognition with a multilingual allophone system.
\newblock In \emph{ICASSP 2020-2020 IEEE International Conference on Acoustics, Speech and Signal Processing (ICASSP)}, pages 8249--8253. IEEE.

\bibitem[{Neubig and Hu(2018)}]{neubig-hu-2018-rapid}
Graham Neubig and Junjie Hu. 2018.
\newblock \href {https://doi.org/10.18653/v1/D18-1103} {Rapid adaptation of neural machine translation to new languages}.
\newblock In \emph{Proceedings of the 2018 Conference on Empirical Methods in Natural Language Processing}, pages 875--880, Brussels, Belgium. Association for Computational Linguistics.

\bibitem[{Neubig et~al.(2019)Neubig, Littell, Chen, Lee, Li, Lin, and Zhang}]{neubig19computeel}
Graham Neubig, Patrick Littell, Chian-Yu Chen, Jean Lee, Zirui Li, Yu-Hsiang Lin, and Yuyan Zhang. 2019.
\newblock \href {https://arxiv.org/abs/1812.05272} {Towards a general-purpose linguistic annotation backend}.
\newblock In \emph{Workshop on The Use of Computational Methods in the Study of Endangered Languages (Compute-EL)}, Honolulu, Hawaii.

\bibitem[{Nguyen et~al.(2021)Nguyen, Jatowt, Coustaty, and Doucet}]{nguyen2021survey}
Thi Tuyet~Hai Nguyen, Adam Jatowt, Mickael Coustaty, and Antoine Doucet. 2021.
\newblock Survey of post-ocr processing approaches.
\newblock \emph{ACM Computing Surveys (CSUR)}, 54(6):1--37.

\bibitem[{{NLLB Team} et~al.(2022){NLLB Team}, Costa-jussà, Cross, Çelebi, Elbayad, Heafield, Heffernan, Kalbassi, Lam, Licht, Maillard, Sun, Wang, Wenzek, Youngblood, Akula, Barrault, Mejia-Gonzalez, Hansanti, Hoffman, Jarrett, Sadagopan, Rowe, Spruit, Tran, Andrews, Ayan, Bhosale, Edunov, Fan, Gao, Goswami, Guzmán, Koehn, Mourachko, Ropers, Saleem, Schwenk, and Wang}]{nllb2022}
{NLLB Team}, Marta~R. Costa-jussà, James Cross, Onur Çelebi, Maha Elbayad, Kenneth Heafield, Kevin Heffernan, Elahe Kalbassi, Janice Lam, Daniel Licht, Jean Maillard, Anna Sun, Skyler Wang, Guillaume Wenzek, Al~Youngblood, Bapi Akula, Loic Barrault, Gabriel Mejia-Gonzalez, Prangthip Hansanti, John Hoffman, Semarley Jarrett, Kaushik~Ram Sadagopan, Dirk Rowe, Shannon Spruit, Chau Tran, Pierre Andrews, Necip~Fazil Ayan, Shruti Bhosale, Sergey Edunov, Angela Fan, Cynthia Gao, Vedanuj Goswami, Francisco Guzmán, Philipp Koehn, Alexandre Mourachko, Christophe Ropers, Safiyyah Saleem, Holger Schwenk, and Jeff Wang. 2022.
\newblock No language left behind: Scaling human-centered machine translation.

\bibitem[{Ogueji et~al.(2021)Ogueji, Zhu, and Lin}]{ogueji-etal-2021-small}
Kelechi Ogueji, Yuxin Zhu, and Jimmy Lin. 2021.
\newblock \href {https://doi.org/10.18653/v1/2021.mrl-1.11} {Small data? no problem! exploring the viability of pretrained multilingual language models for low-resourced languages}.
\newblock In \emph{Proceedings of the 1st Workshop on Multilingual Representation Learning}, pages 116--126, Punta Cana, Dominican Republic. Association for Computational Linguistics.

\bibitem[{Ranathunga et~al.(2023)Ranathunga, Lee, Prifti~Skenduli, Shekhar, Alam, and Kaur}]{ranathunga2023neural}
Surangika Ranathunga, En-Shiun~Annie Lee, Marjana Prifti~Skenduli, Ravi Shekhar, Mehreen Alam, and Rishemjit Kaur. 2023.
\newblock Neural machine translation for low-resource languages: A survey.
\newblock \emph{ACM Computing Surveys}, 55(11):1--37.

\bibitem[{Rijhwani et~al.(2020)Rijhwani, Anastasopoulos, and Neubig}]{rijhwani-etal-2020-ocr}
Shruti Rijhwani, Antonios Anastasopoulos, and Graham Neubig. 2020.
\newblock \href {https://doi.org/10.18653/v1/2020.emnlp-main.478} {{OCR} {P}ost {C}orrection for {E}ndangered {L}anguage {T}exts}.
\newblock In \emph{Proceedings of the 2020 Conference on Empirical Methods in Natural Language Processing (EMNLP)}, pages 5931--5942, Online. Association for Computational Linguistics.

\bibitem[{Team et~al.(2022)Team, Costa-jussà, Cross, Çelebi, Elbayad, Heafield, Heffernan, Kalbassi, Lam, Licht, Maillard, Sun, Wang, Wenzek, Youngblood, Akula, Barrault, Gonzalez, Hansanti, Hoffman, Jarrett, Sadagopan, Rowe, Spruit, Tran, Andrews, Ayan, Bhosale, Edunov, Fan, Gao, Goswami, Guzmán, Koehn, Mourachko, Ropers, Saleem, Schwenk, and Wang}]{nllbteam2022language}
NLLB Team, Marta~R. Costa-jussà, James Cross, Onur Çelebi, Maha Elbayad, Kenneth Heafield, Kevin Heffernan, Elahe Kalbassi, Janice Lam, Daniel Licht, Jean Maillard, Anna Sun, Skyler Wang, Guillaume Wenzek, Al~Youngblood, Bapi Akula, Loic Barrault, Gabriel~Mejia Gonzalez, Prangthip Hansanti, John Hoffman, Semarley Jarrett, Kaushik~Ram Sadagopan, Dirk Rowe, Shannon Spruit, Chau Tran, Pierre Andrews, Necip~Fazil Ayan, Shruti Bhosale, Sergey Edunov, Angela Fan, Cynthia Gao, Vedanuj Goswami, Francisco Guzmán, Philipp Koehn, Alexandre Mourachko, Christophe Ropers, Safiyyah Saleem, Holger Schwenk, and Jeff Wang. 2022.
\newblock \href {http://arxiv.org/abs/2207.04672} {No language left behind: Scaling human-centered machine translation}.

\bibitem[{Viswanathan et~al.(2023)Viswanathan, Zhao, Bertsch, Wu, and Neubig}]{prompt2model}
Vijay Viswanathan, Chenyang Zhao, Amanda Bertsch, Tongshuang Wu, and Graham Neubig. 2023.
\newblock \href {http://arxiv.org/abs/2308.12261} {Prompt2model: Generating deployable models from natural language instructions}.

\bibitem[{Wan et~al.(2018)Wan, Wang, Papir, and Moreno}]{wan2018generalized}
Li~Wan, Quan Wang, Alan Papir, and Ignacio~Lopez Moreno. 2018.
\newblock Generalized end-to-end loss for speaker verification.
\newblock In \emph{2018 IEEE International Conference on Acoustics, Speech and Signal Processing (ICASSP)}, pages 4879--4883. IEEE.

\bibitem[{Wittenburg et~al.(2006)Wittenburg, Brugman, Russel, Klassmann, and Sloetjes}]{wittenburg-etal-2006-elan}
Peter Wittenburg, Hennie Brugman, Albert Russel, Alex Klassmann, and Han Sloetjes. 2006.
\newblock \href {http://www.lrec-conf.org/proceedings/lrec2006/pdf/153_pdf.pdf} {{ELAN}: a professional framework for multimodality research}.
\newblock In \emph{Proceedings of the Fifth International Conference on Language Resources and Evaluation ({LREC}{'}06)}, Genoa, Italy. European Language Resources Association (ELRA).

\bibitem[{Xue et~al.(2022)Xue, Barua, Constant, Al-Rfou, Narang, Kale, Roberts, and Raffel}]{xue-etal-2022-byt5}
Linting Xue, Aditya Barua, Noah Constant, Rami Al-Rfou, Sharan Narang, Mihir Kale, Adam Roberts, and Colin Raffel. 2022.
\newblock \href {https://doi.org/10.1162/tacl_a_00461} {{B}y{T}5: Towards a token-free future with pre-trained byte-to-byte models}.
\newblock \emph{Transactions of the Association for Computational Linguistics}, 10:291--306.

\bibitem[{Yu et~al.(2020)Yu, Kang, Chen, Wu, and Zhao}]{yu2020acoustic}
Chongchong Yu, Meng Kang, Yunbing Chen, Jiajia Wu, and Xia Zhao. 2020.
\newblock Acoustic modeling based on deep learning for low-resource speech recognition: An overview.
\newblock \emph{IEEE Access}, 8:163829--163843.

\end{thebibliography}

\end{document}